\documentclass[10pt]{article}

\usepackage{authblk}
\usepackage{amsmath}
\usepackage{bm}
\usepackage{color}
\usepackage{graphicx}
\usepackage{amssymb}

\newcommand{\prop}[2]{\mathcal{{\rm Prop}}_{#1} \bigl[#2\bigl]}

\DeclareMathOperator*{\argmin}{argmin}

\newcommand*{\argminl}{\argmin\limits}

\title{Autoencoder-based holographic image restoration}
\author[1]{Tomoyoshi Shimobaba\thanks{shimobaba@faculty.chiba-u.jp}}
\author[1]{Yutaka Endo}
\author[1]{Ryuji Hirayama}
\author[1]{Yuki Nagahama}
\author[1]{Takayuki Takahashi}
\author[1]{Takashi Nishitsuji}
\author[1]{Takashi Kakue}
\author[2]{Atsushi Shiraki}
\author[3]{Naoki Takada}
\author[4]{Nobuyuki Masuda}
\author[1]{Tomoyoshi Ito}
\affil[1]{Graduate School of Engineering, Chiba University, 1-33 Yayoi-cho, Inage-ku, Chiba, 263-8522, Japan}
\affil[2]{Institute of Management and Information Technologies, Chiba University, 1-33 Yayoi-cho, Inage-ku, Chiba, 263-8522, Japan}
\affil[3]{Research and Education Faculty, Kochi University, 2-5-1 Akebono-cho, Kochi, Kochi 780-8520, Japan}
\affil[4]{Department of Applied Electronics, Tokyo University of Science,6-3-1 Niijuku, Katsushika-ku, Tokyo 125-8585, Japany}

\begin{document}
\maketitle

\begin{abstract}
We propose a holographic image restoration method using an autoencoder, which is an artificial neural network. Because holographic reconstructed images are often contaminated by direct light, conjugate light, and speckle noise, the discrimination of reconstructed images may be difficult. In this paper, we demonstrate the restoration of  reconstructed images from holograms that record page data in holographic memory and QR codes by using the proposed method. 

\end{abstract}

\section{Introduction}
Digital holography records the amplitude and phase of a real-world object as a hologram using electronic imaging devices and reconstructs the object's amplitude and phase from the holograms on a computer \cite{poon2006digital}.
This allows digital holography to be used in three-dimensional measurements, microscopy, encryption \cite{refregier1995optical}, holographic memory \cite{hong1995volume,matoba2004secure}, and information hiding \cite{aoki2001watermarking,kishk2002information}.

Recently, using digital holography for encryption and information hiding has attracted attention because information security has become increasingly important.
For example, double random-phase encryption \cite{refregier1995optical} encrypts an image by using two random-phase plates (i.e., the encryption key), and this image can be decrypted using the same key. 
Although optical encryption methods such as this one provide high-speed processing and a large number of  encryption keys that are random-phase plates, wavelength, and distance, the decrypted images are often contaminated by speckle noise.
This means that to discriminate the decrypted images.

To avoid these difficulties, a quick response (QR) code is often used instead of raw images or raw information \cite{barrera2013optical, jiao2017qr}. 
A QR code is a two-dimensional bar code comprising binary rectangles.
QR codes reconstructed from holograms improve the retrieval of the information. However, due to the speckle noise, it is still difficult to recognize reconstructed QR codes in certain recording conditions, which consequently makes retrieving the raw information difficult.

Holographic memory \cite{hong1995volume,matoba2004secure} is a type of optical memory.
In holographic memory, the digital data that is to be recorded is converted into a two-dimensional pattern comprising binary rectangles, which are referred to as the page data.
This page data is recorded onto a recording medium, such as a volume hologram. 
Holographic memory has a fast access speed because data can be read and written as two-dimensional page data. Moreover, multiple page data can be recorded in the same recording area because of the multiplex recording characteristics of holograms. 
Furthermore, because holograms can be redundantly recorded, even if the recording medium is slightly missing, the recorded data should be able to restored.
However, similar to optical encryption, page data reconstructed from a holographic memory is often degraded by speckle noise. 
This degradation may make it difficult to recover the recorded digital data.

We therefore propose a holographic image restoration method that uses an autoencoder \cite{hinton2006reducing}.
An autoencoder is an artificial neural network. 
The autoencoder has a function that allows it to restore the desired output, even if the input is degraded by noise.
In this paper, we provide a method to restore reconstructed images from holograms that were are recorded in a manner similar to the page data used in holographic memory and QR codes.

Section 2 describes the proposed restoration method that uses the autoencoder, while section 3 demonstrates the effectiveness of the proposed method using numerical simulation. Section 4 concludes this work. 

\section{Proposed method}
\begin{figure}[htbp]
\centering
\fbox{\includegraphics[width=\linewidth]{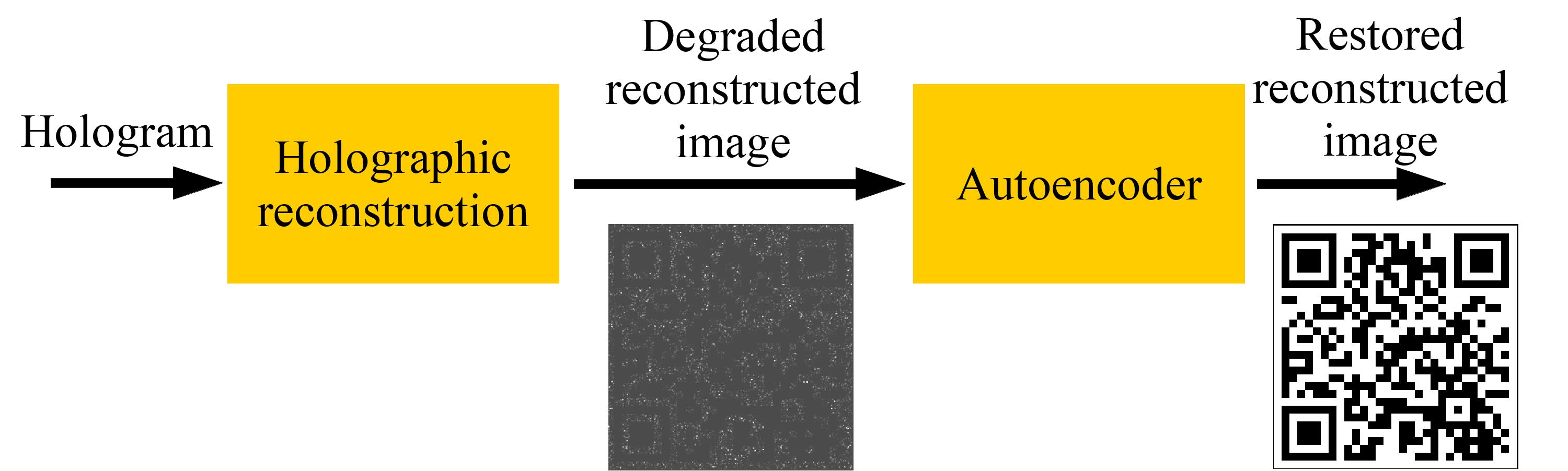}}
\caption{Autoencoder-based holographic image restoration.}
\label{fig:system}
\end{figure}

Figure \ref{fig:system} shows the autoencoder-based holographic image restoration method.
The reconstructed images are obtained by a numerical diffraction calculation or by optical reconstruction from the holograms.
Although the reconstructed images are contaminated by speckle noise, the autoencoder can restore clearer reconstructed images.

In general, autoencoders are used for two purposes: (1) to construct a neural network that can restore original patterns from patterns that have been degraded by noise and (2) to pre-train each layer of a deep neural network. 
In this paper, we use autoencoders as outlined in (1).
in order to train an autoencoder, we need to prepare a large number of training datasets with both original images and the images reconstructed from the holograms that recorded the original images.

We generate amplitude holograms, $I(\bm{m})$, from original images with different random phase using
\begin{equation}
I({\bm m}) = |O({\bm m}) + R({\bm m})|^2,
\end{equation}
where $O({\bm m})$ is the object light and $R({\bm m})$ is the reference light.
${\bm m}$ denotes a two-dimensional position vector in the hologram plane.
The object light is obtained from the original image, $u({\bm n})$, using 
\begin{equation}
O({\bm m}) = \prop{z}{u({\bm n}) \exp(i 2 \pi p({\bm n}))},
\label{eqn:fre2}
\end{equation}
where $\prop{z}{\cdot}$ denotes the diffraction operator with a propagation distance of, $z$, and $p(\bm{n})$ is a pseudo-random number ranging from 0 to 1, which is used for expressing the random phase plates.
${\bm n}$ denotes a two-dimensional position vector in the object plane.
The reconstructed image, $u'(\bm{n})$, from the hologram is obtained by
\begin{equation}
u`(\bm{n}) = |\prop{-z}{I(\bm{m})}|^2.
\label{eqn:fre2}
\end{equation}
The reconstructed image is then degraded by speckle noise, direct light and conjugate light, so that we can restore the original image via an autoencoder.

\begin{figure}[htbp]
\centering
\fbox{\includegraphics[width=\linewidth]{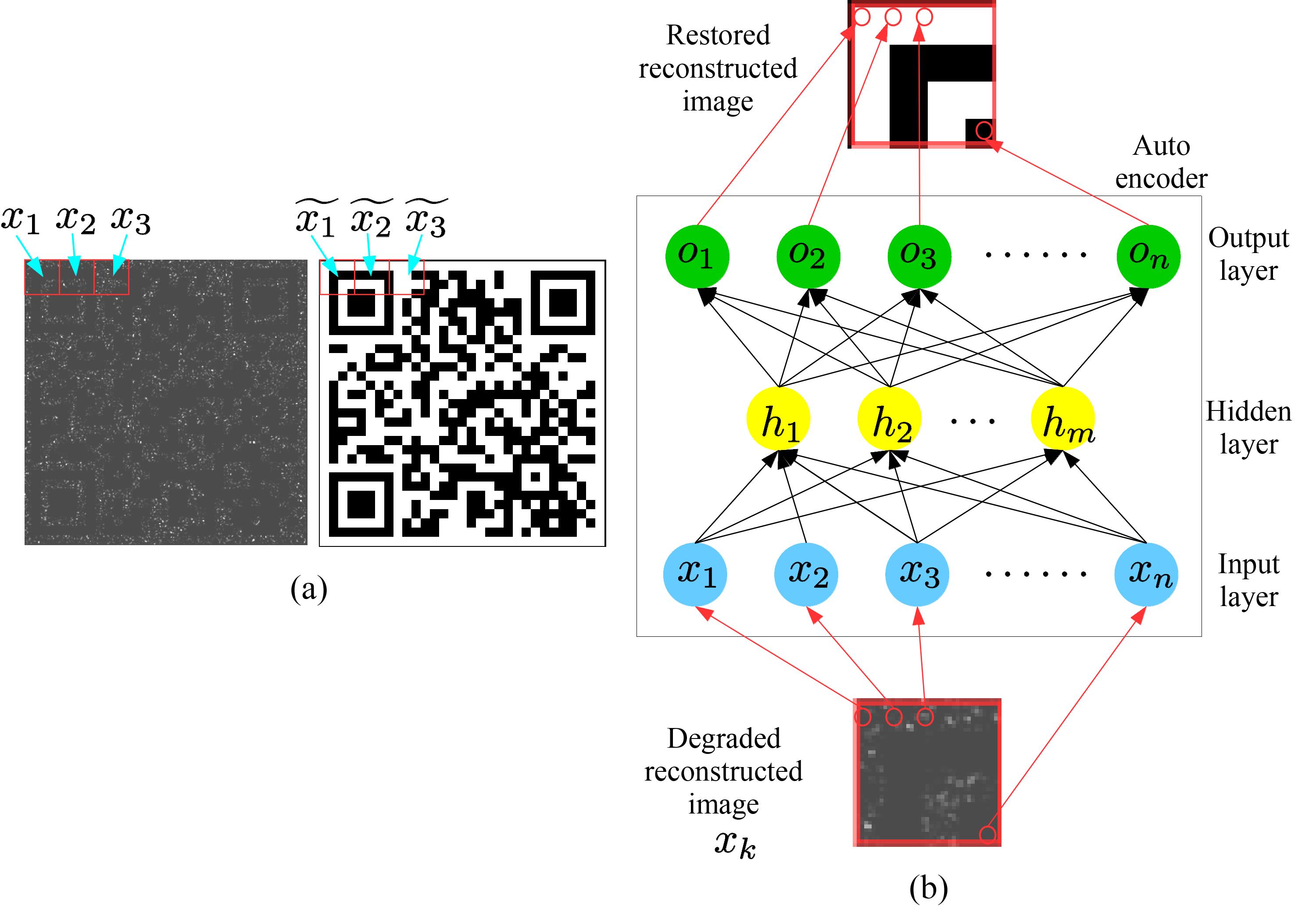}}
\caption{Autoencoder: (a) original images and reconstructed images are divided into sub-patterns, (b) input, hidden and output layers in the autoencoder.}
\label{fig:autoencoder}
\end{figure}

Figure \ref{fig:autoencoder} shows the autoencoder used  in this study; it comprises an input layer, $\bm x$, a hidden layer, $\bm h$ and an output layer, $\bm{o}$, which is used to restore the holographic reconstruction.

The original images and reconstructed images are divided into sub-patterns with $N \times N$ pixels, as shown in Fig. \ref{fig:autoencoder}(a).
Each pixel of the $k$-th subpattern of the reconstructed  and original images is vectorized as
\begin{eqnarray}
\bm{x}_k &=& [u`(\bm{n}_1), u`(\bm{n}_2), \cdots u`(\bm{n}`_{N^2})]^T, \\
\bm{\tilde{x}}_k &=& [u(\bm{n}_1), u(\bm{n}_2), \cdots u(\bm{n}_{N^2})]^T,
\end{eqnarray}
where $\bm{x}_k \in \mathbb{R}^{N^2 \times 1}$ and $\bm{\tilde{x}}_k \in \mathbb{R}^{N^2 \times 1}$.

As shown in Fig.\ref{fig:autoencoder}(b), $\bm{x}_k$ is the input to the input layer, $\bm x$.
The hidden layer, $\bm{h} \in \mathbb{R}^{M \times 1}$, is calculated using
\begin{equation}
\bm{h} = f(\bm{W} \bm{x} + \bm{b}),
\label{eqn:input_hidden}
\end{equation}
where $f(\cdot)$ is an element-wise activation function, $\bm{W} \in \mathbb{R}^{M \times N^2}$ is the parameter matrix between the input and hidden layers, and $\bm{b} \in \mathbb{R}^{M \times 1} $ is the bias vector.
In this paper, we use the rectified linear unit (ReLU) function as the activation function.

The output layer, $\bm{o} \in \mathbb{R}^{N^2 \times 1} $, is calculated by
\begin{equation}
\bm{o} = f(\bm{\widetilde{W}} \bm{h} + \bm{\tilde{b}}),
\label{eqn:hidden_output}
\end{equation}
where $\bm{\widetilde{W}}  \in \mathbb{R}^{N^2 \times M}  $ is the parameter matrix between the hidden and output layers, and $\bm{\tilde{b}}  \in \mathbb{R}^{N^2 \times 1} $ is the bias vector.
Substituting Eq.(\ref{eqn:input_hidden}) into Eq.(\ref{eqn:hidden_output}), we can obtain
\begin{equation}
\bm{o} = f(\bm{\widetilde{W}} f(\bm{W} \bm{x} + \bm{b}) + \bm{\tilde{b}}).
\label{eqn:input_output}
\end{equation}

In the autoencoder's training process, a large number of dataset with the original and reconstructed images is used to find the parameters ($\bm{W}$ and $\bm{\widetilde{W}}$) and biases ($\bm{b}$ and $\bm{\tilde{b}}$) that can minimize the loss function, $e$ (least-squares error), for all of $\bm{\tilde{x}}_k$ and $\bm o$.
The loss function is defined as
\begin{equation}
e = | \bm{\tilde{x}} - f(\bm{\widetilde{W}} f(\bm{W} \bm{x} + \bm{b}) + \bm{\tilde{b}}) |^2. 
\label{eqn:loss}
\end{equation}
Therefore, we optimize the following equation
\begin{equation}
\argminl_{\bm{W},\bm{\widetilde{W}},\bm{b},\bm{\tilde{b}}} \sum_k e. 
\label{eqn:min}
\end{equation}

We used Adam \cite{kingma2014adam}, which is a stochastic gradient descent (SGD) method, to minimize Eq.(\ref{eqn:min}). 
This SGD randomly selects $B$ datasets from all of the datasets $\bm{{x}}_k$ and $\bm{\tilde{x}}_k$.
$B$ is referred to as the batch size, and we used a batch size of 100.
In addition, we used Dropout method \cite{srivastava2014dropout} to prevent overfitting in the autoencoder.
Dropout randomly disables $N_d$ percent of units during the auto encoder the training process;  we used $N_d=0.8$\%. 

After the learning process was complete, degraded reconstructed images, $\bm{{x}}_k$, were put into the autoencoder; this enabled us to obtain the restored images.

\section{Results}
\begin{figure}[htbp]
\centering
\fbox{\includegraphics[width=\linewidth]{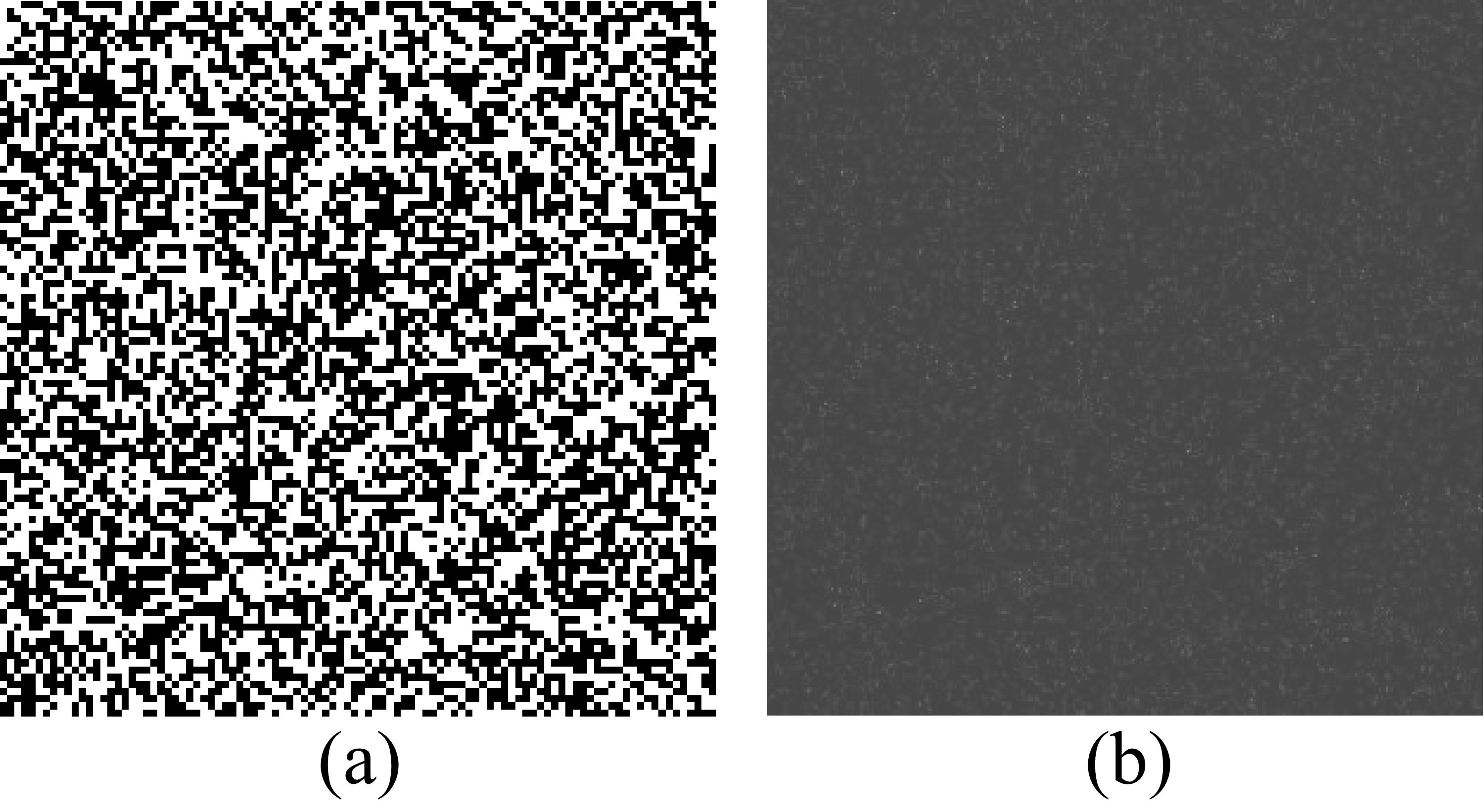}}
\caption{Example of page data: (a) original page data and (b) reconstructed page data.}
\label{fig:fig_page_data}
\end{figure}

We we will begin this section by showing the restoration of the page data used in holographic memory.
Figure \ref{fig:fig_page_data} shows an example of the page data and the reconstructed image obtained from the hologram using the diffraction calculation.
The size of the white and black rectangles in the page data is $10 \times 10$ pixels.
The simulation conditions are shown in Table \ref{tbl:condition}.
We used Chainer \cite{tokui2015chainer} as a deep learning framework and used the CWO++ library \cite{shimobaba2012computational} for the generation and reconstruction of the holograms.
The reconstructed image in Fig.\ref{fig:fig_page_data} is contrast enhanced because the contrast was dark. This darkness was due to a high amount of speckle noise contamination. Raw reconstructed images were used in  in the training process of the autoencoder.

\begin{table}[]
\centering
\caption{Calculation conditions.}
\label{tbl:condition}
\begin{tabular}{|l|l|lll}
\cline{1-2}
Number of pixels & 1,000 $\times$ 1,000 & & & \\ \cline{1-2}
Propagation distance, $z$ & 0.05 m & & & \\ \cline{1-2}
Sampling pitch, $p$ & 4 $\mu$m & & & \\ \cline{1-2}
Wavelength, $\lambda$ & 633 nm & & & \\ \cline{1-2}
\end{tabular}
\end{table}

A restored image is obtained from the degraded reconstructed image via the autoencoder.
We divide a $1,000 \times 1,000$-pixel page data into $50 \times 50$ subpatterns comprising $20 \times 20$ pixels. 
The reconstructed images are similarly divided into $50 \times 50$ subpatterns with $20 \times 20$ pixels. 
The subpatterns of the page data and reconstructed images are vectorized to $\bm{\tilde{x}}$ and $\bm{x}$ with 400 elements, respectively. Therefore, the number of units in the input and output layers is 400.
The number of units in the hidden layer is set to 50, which was empirically decided upon.

\begin{figure}[htbp]
\centering
\fbox{\includegraphics[width=\linewidth]{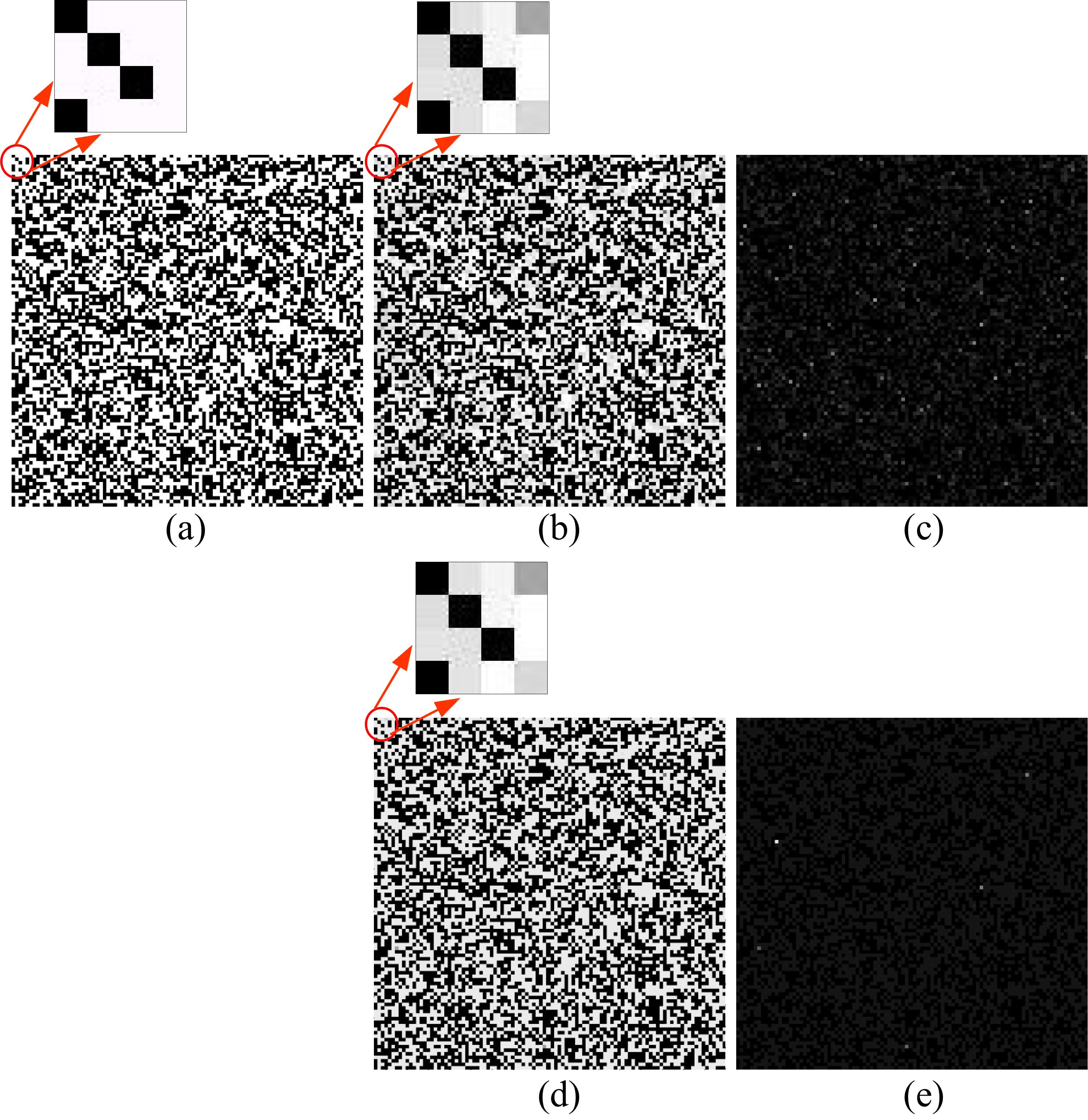}}
\caption{Restoration of the page data: (a) original page data, (b) page data restored using the autoencoder (five iterations), (c) difference between images of (a) and (b), (d) page data restored using the autoencoder (40 iterations), and (e) difference between images of (a) and (d).}
\label{fig:page_data_47500}
\end{figure}

\begin{figure}[htbp]
\centering
\fbox{\includegraphics[width=\linewidth]{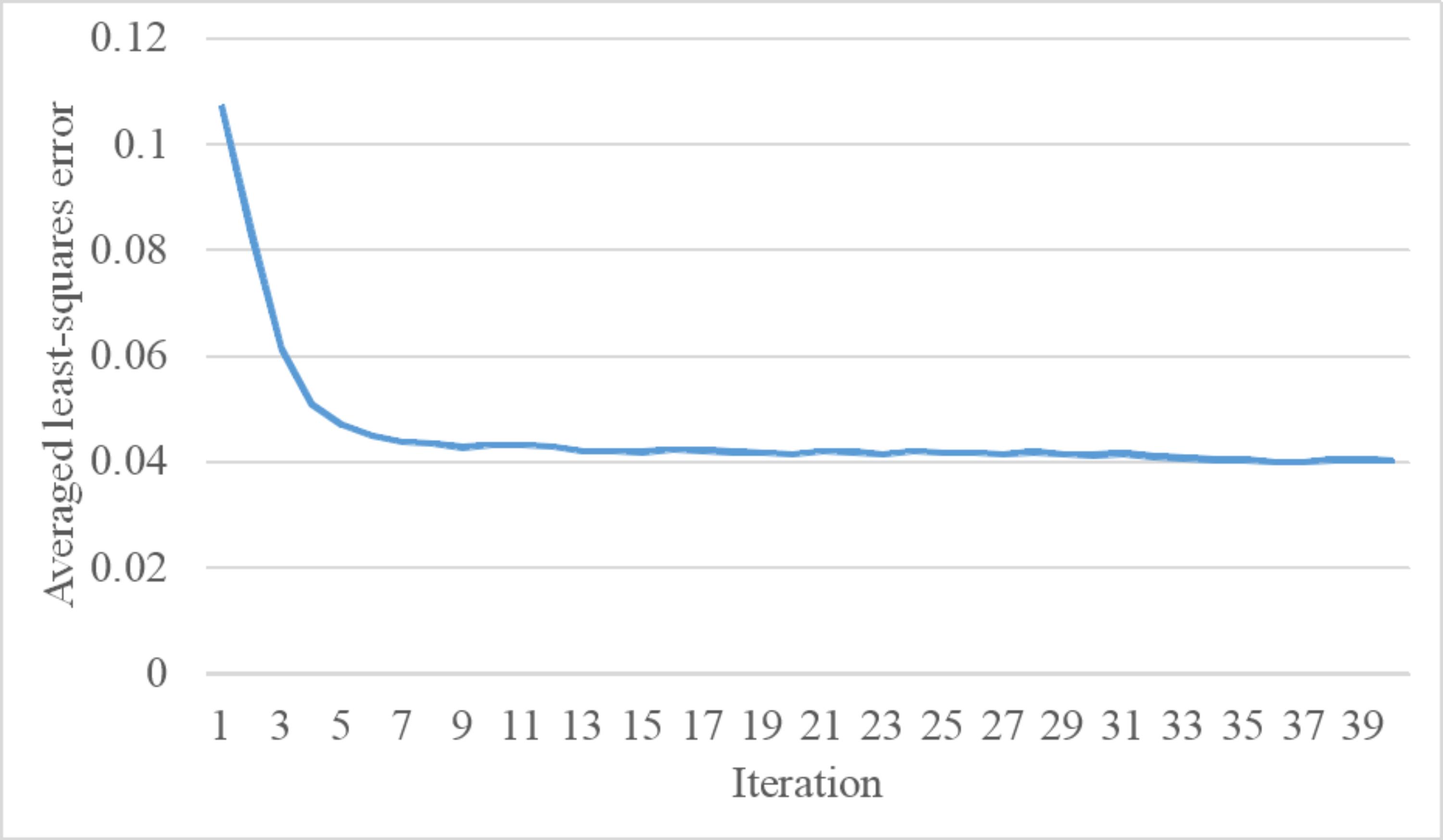}}
\caption{Averaged least-squares error as a function of the number of SGD iterations.}
\label{fig:graph}
\end{figure}

Figure \ref{fig:page_data_47500} shows the images that had been restored by the autoencoder, which had been trained by a dataset that comprised 47,500 subpatterns of  page data and reconstructed images.
Figure \ref{fig:page_data_47500}(a) is the original page data, while Fig. \ref{fig:page_data_47500}(b) is the page data restored by the autoencoder after it had been  trained by five iterations of the SGD during the training process.
Figure \ref{fig:page_data_47500}(c) displays the difference between the images in Figs.\ref{fig:page_data_47500}(a) and \ref{fig:page_data_47500}(b); as can be seen, there is quite a large difference.
Figure \ref{fig:page_data_47500}(d) is the page data restored by the autoencoder after it had been trained by 40 iterations of the SGD during the training process.
Figure \ref{fig:page_data_47500}(e) is the difference between the images in Figs.\ref{fig:page_data_47500}(a) and \ref{fig:page_data_47500}(d).
We can see that now there is less of a difference between two; thus, a better restored image has been obtained for Fig.\ref{fig:page_data_47500}(d) than the one shown in Fig.\ref{fig:page_data_47500}(b).
Figure \ref{fig:graph} is a graph of the averaged least-squares error based on Eq.(\ref{eqn:loss}) as a function of the number of SGD iterations.

\begin{figure}[htbp]
\centering
\fbox{\includegraphics[width=\linewidth]{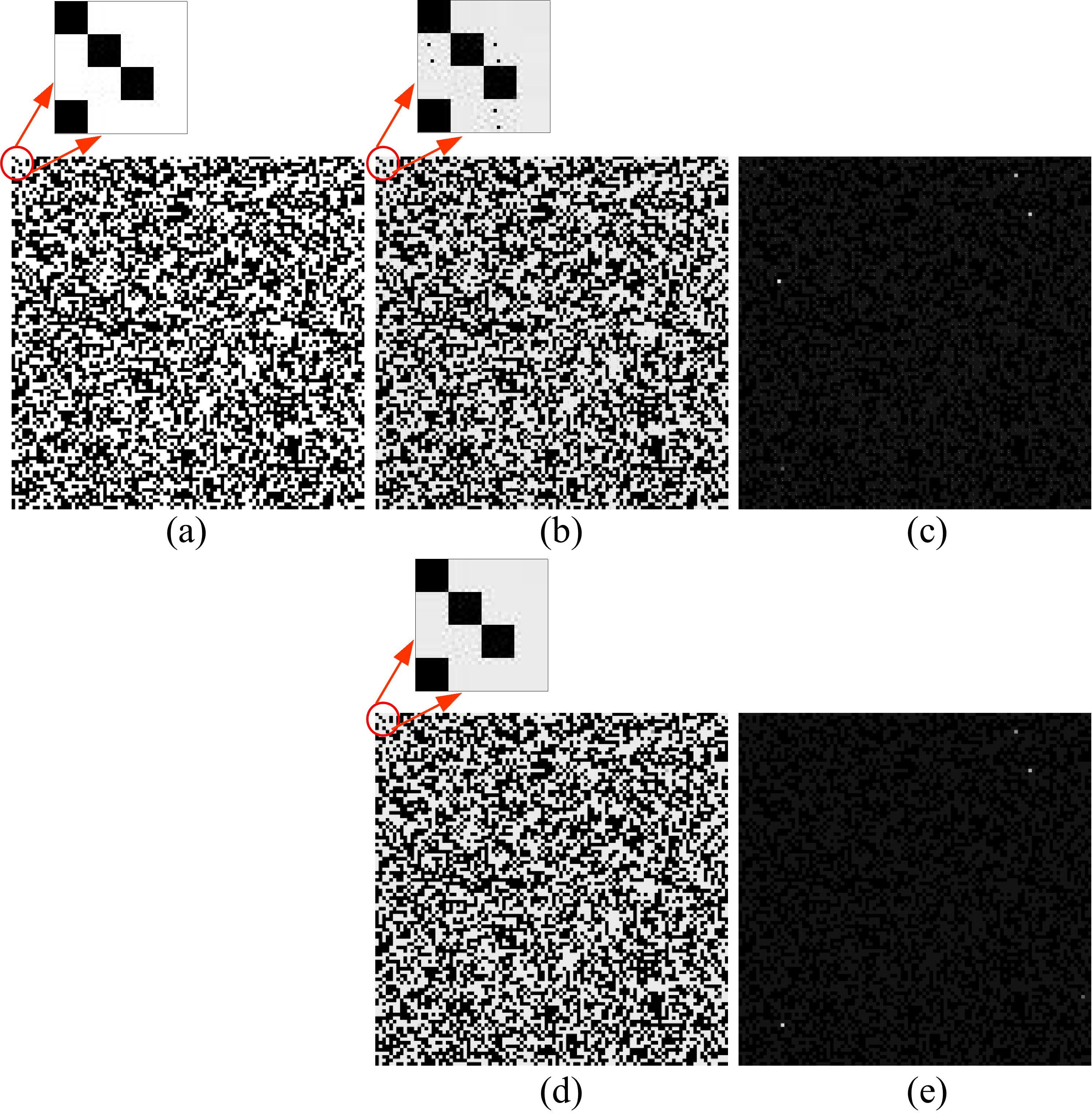}}
\caption{Restoration of the page data: (a) original page data, (b) page data restored using the autoencoder (five iterations), (c) difference between images of (a) and (b), (d) page data restored using the autoencoder (40 iterations), and (e) difference between the image of (a) and (d).}
\label{fig:page_data_247500}
\end{figure}

To obtain an even better restoration, we trained the autoencoder using more datasets.
Figure \ref{fig:page_data_247500} shows the restored image using an autoencoder that had been trained by a dataset comprising 247,500 subpatterns of the page data and reconstructed images.
Figure \ref{fig:page_data_247500}(a) is the original page data, while Fig. \ref{fig:page_data_247500}(b) is the page data restored by the autoencoder after it had been trained by five iterations of the SGD during the training process.
Figure \ref{fig:page_data_247500}(c) is the difference  between the images in Figs.\ref{fig:page_data_247500}(a) and \ref{fig:page_data_247500}(b).
The number of differences between them has decreased, and the restored image has become more uniform in  comparison to Fig.\ref{fig:page_data_47500}(b). However,  there are still some missing pixels in Fig.\ref{fig:page_data_247500}(b).
Figure \ref{fig:page_data_247500}(d) is the page data restored by the autoencoder after it had been trained by 40 iterations of the SGD during the training process.
Figure \ref{fig:page_data_247500}(e) is the difference  between the images in Figs.\ref{fig:page_data_247500}(a) and \ref{fig:page_data_247500}(d).
We can see from Fig.\ref{fig:page_data_247500}(e) that there are now fewer differences and missing pixels between the two images; as such, we can say that a better restored image has been obtained than for the image shown in Fig.\ref{fig:page_data_247500}(b).

\begin{figure}[htbp]
\centering
\fbox{\includegraphics[width=\linewidth]{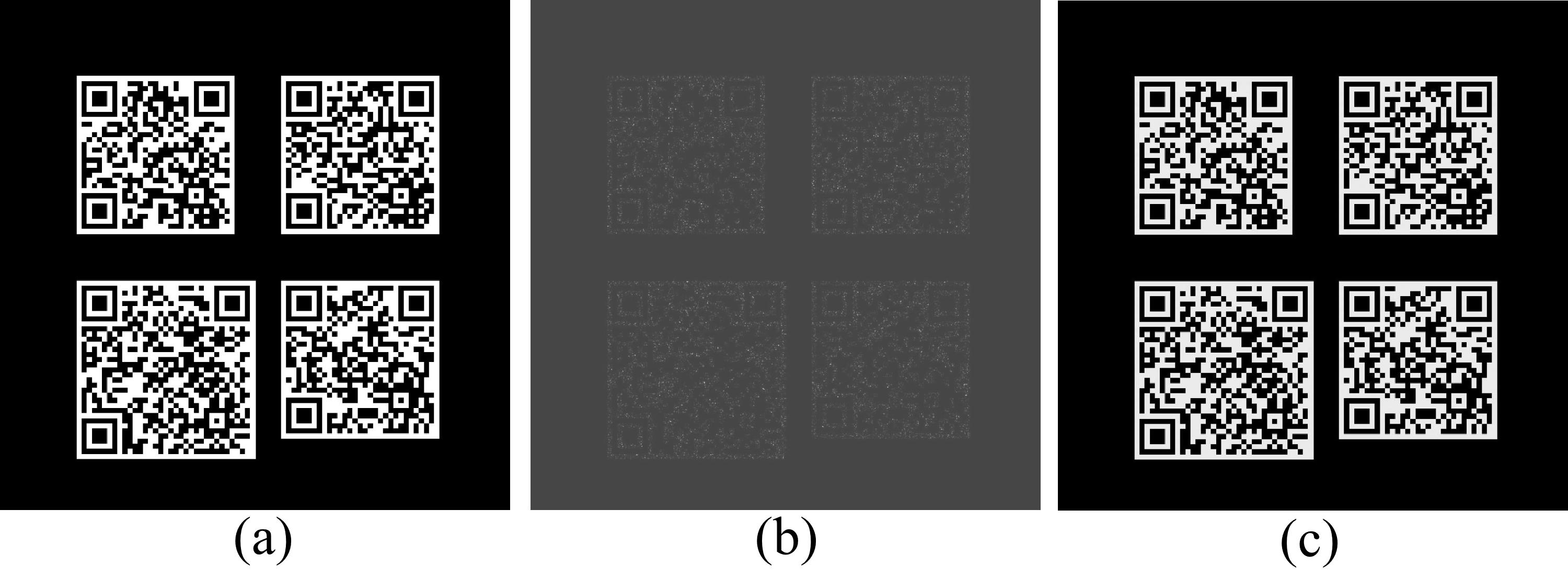}}
\caption{Restoration of QR codes: (a) original QR codes, (b) reconstructed QR codes, and (c) restored QR codes via the autoencoder.}
\label{fig:qr}
\end{figure}

Figure \ref{fig:qr} shows the restoration of QR codes. 
We used the same calculation condition and training dataset that were used to obtain the images in Fig.\ref{fig:page_data_47500}.
Figures \ref{fig:qr}(a), \ref{fig:qr}(b) and \ref{fig:qr}(c) are the original QR codes, the QR codes reconstructed using the diffraction calculation, and the QR codes restored using the autoencoder, respectively.
The original QR codes contained four different uniform resource locators (URLs).
The reconstructed QR codes in Fig.\ref{fig:qr}(b) are contrast enhanced. However, we were unable to  recognize all of the reconstructed QR codes using the cameras on our smart phones; this was because of the large amount of speckle noise in the images.
Conversely, we were able to readily recognize all of the restored QR codes using our smart phones.

\section{Conclusion}
In this paper, we have presented a proposed  autoencoder-based holographic image restoration method. 
The demonstration of the restorations from the reconstructed images of the page data and QR codes  shows the effectiveness of our proposed method.
The reconstructed images of the QR codes directly from the diffraction calculation could not be recognized due to the large amount of speckle noise, whereas the QR codes restored using the autoencoder could be correctly recognized by our smartphones.
In future, we will look to further improve restoration quality by using deep autoencoders, convolution autoencoders, and deep convolutional neural networks. In addition, we will use this technique to perform super-resolution and restore stacked page data and QR codes.

\section*{Funding Information}

\textbf{Funding.} 
This work was partially supported by JSPS KAKENHI Grant Numbers 16K00151 and 25240015. 

\bibliographystyle{unsrt}
\bibliography{sample}

\end{document}